\documentclass[conference]{IEEEtran}
\IEEEoverridecommandlockouts
\usepackage{cite}
\usepackage{url}
\usepackage{amsmath,amssymb,amsfonts}
\usepackage{algorithmic}
\usepackage{graphicx}
\usepackage{textcomp}
\usepackage{xcolor}
\usepackage{multirow}
\def\BibTeX{{\rm B\kern-.05em{\sc i\kern-.025em b}\kern-.08em
    T\kern-.1667em\lower.7ex\hbox{E}\kern-.125emX}}

\input{mysymbol.sty}

\begin{document}

% \title{Overview and Experimental Evaluation of Hyperbolic Graph Representation Learning Methods\thanks{This work was partially funded by CSIC i+d project ``Geometr\'ia en Redes Complejas y Aplicaciones a Aprendizaje Autom\'atico''.}}
% \title{Hyperbolic Graph Representation Learning: Experimental Evaluation and an Open-Source Framework\thanks{This work was partially funded by CSIC i+d project ``Geometr\'ia en Redes Complejas y Aplicaciones a Aprendizaje Autom\'atico''.}}
% \title{Hyperbolic Graph Representation Learning: Open-Source Framework and Experimental Evaluation\thanks{This work was partially funded by CSIC i+d project ``Geometr\'ia en Redes Complejas y Aplicaciones a Aprendizaje Autom\'atico''.}}
\title{A Unified Framework of Hyperbolic Graph Representation Learning Methods\thanks{This work was partially funded by CSIC i+d project ``Geometr\'ia en Redes Complejas y Aplicaciones a Aprendizaje Autom\'atico''.}}

% \title{Hyperbolic Graph Representation Learning: Overview, Open-Source Framework and Experimental Evaluation\thanks{This work was partially funded by CSIC i+d project ``Geometr\'ia en Redes Complejas y Aplicaciones a Aprendizaje Autom\'atico''.}}

% \and
% \and
% \and
% \and
% \and
\author{\IEEEauthorblockN{
Sof\'ia P\'erez Casulo\IEEEauthorrefmark{1},
Marcelo Fiori\IEEEauthorrefmark{1}\IEEEauthorrefmark{2},
Bernardo Marenco\IEEEauthorrefmark{1}\IEEEauthorrefmark{2} and 
Federico Larroca\IEEEauthorrefmark{1}\IEEEauthorrefmark{2}}
\IEEEauthorblockA{\IEEEauthorrefmark{1}Facultad de Ingenier\'ia, Universidad de la Rep\'ublica, Uruguay\\
\IEEEauthorrefmark{2} Centro Interdisciplinario en Ciencia de Datos y Aprendizaje Automático (CICADA), Universidad de la Rep\'ublica, Uruguay\\
Emails: \{sperez,mfiori,bmarenco,flarroca\}@fing.edu.uy}
}

\maketitle

\begin{abstract}
Hyperbolic geometry has emerged as an effective latent space for representing complex networks, owing to its ability to capture hierarchical organization and heterogeneous connectivity patterns using low-dimensional embeddings. As a result, numerous hyperbolic graph representation learning methods have been proposed in recent years. However, their practical adoption and systematic comparison remain challenging, as implementations are fragmented and shared tools for reproducible and fair evaluation are lacking.
%However, their practical adoption and systematic comparison remain challenging due to fragmented implementations, and the community lacks shared tools to enable reproducible and comparable evaluations.
%heterogeneous optimization pipelines and limited support for reproducible evaluation.
% 
In this work, we introduce a unified open-source framework for hyperbolic graph representation learning that integrates several widely used embedding methods under a common optimization interface. The novel framework enables consistent training, visualization, and evaluation of hyperbolic embeddings, and interfaces seamlessly with standard network analysis tools. Leveraging this unified setup, we conduct an experimental study of hyperbolic embedding methods on real-world networks, focusing on two canonical downstream tasks: link prediction and node classification.
% 
% Our results provide a comparative assessment of the considered methods and highlight how \red{cambiar de acá en adelante cunado tengamos los resultados} modeling choices, embedding dimensionality, and geometric assumptions influence performance across tasks. 
Beyond predictive accuracy, the study offers practical insights into the strengths and limitations of existing approaches, thereby facilitating informed method selection and fostering reproducible research in hyperbolic graph representation learning.
\end{abstract}

\begin{IEEEkeywords}
Graph representation learning, Hyperbolic geometry, Latent space graph models, Reproducible research 
\end{IEEEkeywords}

\section{Introduction}

% \red{

% Embeddings geométricos para grafos es una excelente idea. 

% De un tiempo a esta parte hay un ``mounting evidence'' de que la geometría inherente es hiperbólica: ventajas y desventajas. 

% Naturalmente han habido unos cuántos métodos propuestos para hallar estos embeddings, pero no están claras las ventajas y desventajas de cada uno.

% Introducing \texttt{HypeGRL}: en un único framework muchos de estos métodos. 

% Ejemplo de uso: desempeño de cada método en dos tareas ``canónicas'': link prediction y node classification.

% El ganador: (supongo que) hydra+.
% }

% Graph Representation Learning (GRL), sometimes called Network Geometry, is the problem of extracting useful low-dimensional representations of the nodes of an observed graph. That is to say, we assume that each node may be represented by an (unobserved) vector in a certain space, where proximity of two nodes in this space typically also represent proximity in the graph (e.g., few number of hops) or a high probability of connection between them, and we would like to estimate the nodes' vectors. Applications of these methods are varied, including visualization, clustering, link prediction and graph machine learning (where these vectors are called node embeddings).

Graph Representation Learning (GRL) deals with the problem of extracting useful low-dimensional representations of the nodes of an observed graph~\cite{hamilton2020graph}. These approaches are commonly based on latent space models, in which each node is assumed to be associated with an unobserved vector in an underlying geometric space~\cite{boguna2021network}. Proximity between node representations is then designed to reflect structural closeness in the graph, such as a small number of hops or a high probability of connection. Estimating these latent vectors enables a wide range of applications, including visualization, clustering, link prediction, and graph machine learning, where such representations are typically referred to as node embeddings~\cite{chami2022machine}.

% In this context, one of the first decisions the practitioner has to take, sometimes without much consideration, is the actual underlying geometry where these vectors live in. Typically, the choice goes to Euclidean geometry, with a proximity notion that is not necessarily the distance; e.g., the inner product for the popular Random Dot Product Graph (RDPG) model. However, a mounting empirical and theoretical evidence suggest that hyperbolic geometry is actually more fitted to represent complex real-life networks. As we discuss with more detail in Sec.\ \ref{sec:network_geometry_hyperbolic}, the negative curvature of this space naturally induces typically observed properties (e.g., heavy-tailed degree distributions), all the while being more parsimonious than its Euclidean counterpart (i.e.\ necessitating a smaller dimension for the vectors to obtain a comparable performance).

In this context, a crucial yet often under-emphasized design choice is the geometry of the latent space in which node representations are embedded. Euclidean geometry is by far the most commonly adopted alternative, with proximity typically defined through distances or inner products, as in the popular Random Dot Product Graph (RDPG) model~\cite{athreya2018rdpg}. However, mounting empirical and theoretical evidence suggests that hyperbolic geometry provides a more faithful model for many complex real-world networks~\cite{krioukov2010hyperbolic,CHEN2025hyperbolic,WANG2025hyperbolic}. As discussed in more detail in Sec.\ \ref{sec:network_geometry_hyperbolic}, the negative curvature of hyperbolic spaces naturally induces structural properties frequently observed in practice, such as hierarchical organization and heavy-tailed degree distributions, while often achieving comparable or superior performance using lower-dimensional representations than their Euclidean counterparts.
% Naturally, several methods have been proposed to estimate the embeddings under the assumption of a hyperbolic geometry. However, the choice of which method to use is complicated by the fact that, when available, implementations use different platforms or programming languages, without a unified framework that encompasses all (or at least most) of the existing methods. This has led to very few studies that compare the relative merits and performance of each of the existing algorithms.

Motivated by these advantages, a variety of methods have been proposed to learn node embeddings under hyperbolic geometry~\cite{garcia2019mercator,jankowski2023dmercator,kitsak2020link,klimovskaia2020poincare,keller2020hydra,papadopoulos2015network,papadopoulos2015network2,nickel2017poincare,nickel2018learning}. Nevertheless, choosing among existing approaches remains challenging in practice. 
% Available implementations, when they exist, 
Existing (and often limited) implementations are developed in different programming languages or platforms, rely on heterogeneous optimization strategies, and expose incompatible interfaces. As a result, desired reproducibility is hindered and fair comparisons between hyperbolic embedding methods are scarce, limiting both empirical understanding and practical adoption.

% Our main contribution is \texttt{HypeGRL}, a Python-based open-source framework that implements the most popular hyperbolic graph representation methods under a unified API, including several visualization and analysis tools. By interfacing with standard network analysis libraries (e.g., NetworkX), our objective is to simplify and encourage the adoption of hyperbolic geometry for graph embedding tasks. The code as well as all the experiments presented here are available at \url{https://github.com/CicadaUY/hyperbolic-embeddings/}. \red{cambiar para la URL nueva si nos mudamos}.

% Our main contribution is \texttt{HypeGRL}, an open-source Python framework that unifies the implementation of several widely used hyperbolic graph representation learning methods under a common API. 
% \red{Sofi, fijate qué te parece esta oración, sobre todo lo de consistent optimization piplines. yo lo veo por el lado de la API digamos, como las llamadas a train, pero vos dirás si conviene decirlo de otra forma. Idem en el abstract.} \green{No se si me convence usar API, va sugerencia: }
{Our main contribution is \texttt{HypeGRL}, an open-source Python framework that brings together multiple hyperbolic GRL methods under a unified training and evaluation environment.}
As we discuss in Sec.\ \ref{sec:network_geometry_hyperbolic}, the framework provides consistent optimization pipelines, visualization utilities and evaluation tools, and interfaces seamlessly with standard network analysis libraries such as NetworkX. By offering a unified and extensible environment, \texttt{HypeGRL} lowers the barrier to the adoption of hyperbolic geometry in graph learning tasks and enables reproducible and systematic experimental analysis. The source code and all experimental scripts are publicly available at \url{https://github.com/CicadaUY/hypeGRL}.

% Equipped with this framework, as a second contribution we present an experimental evaluation of all these embedding methods. To measure their performance we evaluate them in two canonical machine learning tasks: link prediction and node classification. Our results, reported in Sec.\ \ref{sec:experiments}, indicate \red{completar}.

Building upon \texttt{HypeGRL}, our second contribution is an experimental evaluation of the chosen hyperbolic embedding methods. We assess their performance on two canonical downstream tasks: link prediction and node classification, using several real-world networks. Beyond reporting predictive accuracy, our analysis highlights systematic differences between methods in terms of computation cost, representation efficiency, and task-dependent performance. Our results in Sec.~\ref{sec:experiments} provide practical insights into the relative merits and limitations of existing hyperbolic embedding approaches.

\section{Hyperbolic Geometry, Embeddings and \texttt{HypeGRL}}\label{sec:network_geometry_hyperbolic}

% \red{un poco fuerte el título, no?}

\subsection{Hyperbolic Geometry}\label{subsec:hiperbolic_geometry}
Hyperbolic geometry provides a natural continuous framework for representing hierarchical and tree-like structures. 
In contrast to Euclidean geometry, hyperbolic space exhibits exponential volume growth with respect to radius, allowing distances to naturally reflect hierarchical depth and branching factors. 
As a consequence, negatively curved spaces admit low-distortion embeddings of trees and tree-like graphs that necessarily incur large distortion in Euclidean geometry~\cite{needham2021visual}.

We consider the $n$-dimensional hyperbolic space $\mathbb{H}^n$, a complete, simply connected Riemannian manifold of constant negative curvature. 
Although $\mathbb{H}^n$ is unique up to isometry, it admits several equivalent coordinate representations (or \emph{models}), each highlighting different geometric or analytic properties. 
We briefly review the most common models used in GRL and their relevance for theoretical analysis and algorithmic constructions.

\noindent\textbf{Lorentz (hyperboloid) model. }
We primarily adopt the Lorentz, or hyperboloid, model of hyperbolic space. Let $\reals^{n+1}$ be endowed with the Minkowski bilinear form
\begin{equation}\label{eq:minkowski}
\langle \bbu,\bbv\rangle
=
\sum_{i=1}^n u_i v_i - u_{n+1} v_{n+1}.
\end{equation}
The $n$-dimensional hyperbolic space $\mathbb{H}^n$ is then realized as
\begin{equation}
\mathbb{H}^n
=
\left\{
\bbu \in 
\reals^{n+1} :
\langle \bbu,\bbu\rangle = -1,\;
u_{n+1}>0
\right\},
\end{equation}
with geodesic distance given by 
\begin{gather}\label{eq:hyperbolic_distance_arcosh}
    d_H(\bbu,\bbv) = \operatorname{arcosh}\bigl(-\langle\bbu,\bbv\rangle\bigr).
\end{gather} 

It is common to write $\bbu=(\mathbf{x},t)$, where the last coordinate plays the role of a time-like component, satisfying $\|\mathbf{x}\|^2 - t^2=-1$ with $t>0$. Geometrically, $\mathbb{H}^n$ corresponds to the upper sheet of a two-sheeted hyperboloid in $\reals^{n+1}$, and its geodesics are given by the intersections of the hyperboloid with Euclidean planes passing through the origin. 

The Lorentz model admits simple closed-form expressions for distances and gradients and provides a convenient global parametrization of hyperbolic space, making it particularly well suited for both theoretical analysis and optimization. This motivates its use as the default representation in \texttt{HypeGRL}. Moreover, several alternative representations of hyperbolic space can be naturally interpreted in terms of the Lorentz model.

For instance, the \textbf{native representation} of hyperbolic space is the polar parametrization in terms of a radial coordinate $r$ (hyperbolic distance to the point at $\bbx=\bb0$ and $t=1$) and angular coordinates $\theta=(\theta_1,\theta_2,\ldots,\theta_{n-1})$. This means that for $n=2$ the distance between two points is
\begin{equation*}
d(\bbu,\bbv)
=
\operatorname{arcosh}\!\left(
\cosh r_u \cosh r_v
-
% \sinh r_u \sinh r_v \cos(\theta_u-\theta_v)
\sinh r_u \sinh r_v \cos(\theta_{u,v})
\right).
\end{equation*}
For large radii, this distance admits the asymptotic approximation
$
d(\bbu,\bbv)
\approx
r_u + r_v + 2\log(\theta_{u,v}/2),
$
which makes explicit the separation between radial growth, capturing hierarchical depth, and angular separation, capturing similarity. This decomposition underlies random hyperbolic graph models and the embedding methods built upon them.

% Since coordinate $u_{n+1}$ is typically associated with time, vectors are sometimes expressed as $\bbu=(\mathbf{x},t)$ with $\|\mathbf{x}\|^2 - t^2=-1$ and $t>0$. In any case, the space corresponds to the upper sheet of a hyperboloid in $\reals^{n+1}$. However, given the metric, geodesics are the intersections of the hyperboloid with Euclidean planes through the origin. 
% This model provides a globally geodesic coordinate system and admits simple closed-form expressions for distances and gradients, making it particularly convenient for both theoretical analysis and optimization. \red{For this reason, \texttt{HypeGRL} internally represents embeddings in the Lorentz model, while supporting conversion to other representations when needed. \textbf{VERIFICAR}. } Furthermore, the rest of the representations are arguably easier to understand as projections of the Lorentz model.

\noindent\textbf{Poincar\'e ball model. }
% The Poincar\'e disk model represents $\mathbb{H}^n$ as the unit ball
% \begin{equation*}
% S^{n-1} = \{\bby\in
% \reals^n : \|\bby\|<1\},
% \end{equation*}
% obtained by stereographic projection of the hyperboloid. The mapping from Lorentz coordinates $(\mathbf{x},t)$ is
% \begin{equation*}
% \bby = \frac{\mathbf{x}}{1+t}.
% \end{equation*}
% This model is conformal: angles are preserved and geodesics appear as circular arcs orthogonal to the boundary. 
% While distances become distorted near the boundary, this representation is useful for geometric intuition and visualization.
The Poincar\'e (or Poincar\'e-Beltrami) ball model is obtained by stereographically projecting the hyperboloid onto the plane $t=0$, using the south pole (located at $\bbx=\bb0$ and $t=-1$) as the center of projection. This construction yields a representation of $\mathbb{H}^n$ as the unit ball
\begin{equation}
\mathbb{B}^{n} = \{\bby\in
\reals^n : \|\bby\|<1\},
\end{equation}
with the mapping from Lorentz coordinates $(\mathbf{x},t)$ given by $\bby = \frac{\mathbf{x}}{1+t}$. The Poincar\'e model is conformal, meaning that angles are preserved, and geodesics appear as circular arcs orthogonal to the boundary of the unit ball. Although distances become increasingly distorted near the boundary, this representation is particularly useful for geometric intuition and visualization.

\subsection{\texttt{HypeGRL}: Hyperbolic Graph Representation Learning}\label{sec:biblioteca_and_embeddings}

We introduce \texttt{HypeGRL}, a Python library that brings together multiple hyperbolic embedding methods within a single unified interface. The library provides a common wrapper for seven widely used algorithms which we briefly present below. %: Poincar\'e Embeddings, Lorentz Embeddings, Poincar\'e Maps, D-Mercator, Hydra, Hydra+ and HyperMap. 
It enables consistent training, evaluation, and visualization across all these methods. Users can seamlessly switch between embedding techniques, allowing for direct and fair comparisons of model performance and geometric behavior, and easily extend the library by including further methods.

Beyond unifying optimization routines, \texttt{HypeGRL} includes tools for converting embeddings across the hyperbolic models we discussed before, among other. These utilities enable flexible visualization and make it possible to analyze embeddings in the coordinate system most appropriate for a given task.

\texttt{HypeGRL} unifies previously fragmented implementations into a single extensible framework, making experimentation with hyperbolic methods easier, benchmarking more consistent, and comparisons across models fully reproducible. The included methods are discussed in what follows.

% The embedding methods considered in this paper are widely used and are based on different hypotheses or pursue different objectives.
\noindent \textbf{Hydra and Hydra+~\cite{keller2020hydra}.}
Assume that pairwise dissimilarities between nodes of a graph are available in the form of matrix $\bbD\in\reals^{N\times N}$; e.g., given by shortest-path distances. The objective of the Hydra method is to find embeddings $\bbu_i \in \mathbb{H}^n$ ($i=1,\ldots,N$) such that the resulting hyperbolic distances approximate the observed distances $D_{i,j}$, by minimizing $\sum_{i,j}|d_H(\bbu_i,\bbu_j)-D_{i,j}|^2$.

Directly minimizing this objective leads to a high-dimensional, non-convex optimization problem that is difficult to solve efficiently. Instead, the authors of~\cite{keller2020hydra} reformulate the problem in the Lorentz model by applying the hyperbolic cosine to the target distances, and using \eqref{eq:hyperbolic_distance_arcosh} they solve the surrogate problem $\sum_{i,j}\|-\langle\bbu_i,\bbu_j\rangle-\operatorname{cosh}(D_{i,j})\|^2$. 
Given the form of \eqref{eq:minkowski}, this minimization is efficiently solved using spectral decompositions. 
The Hydra+ variant subsequently refines this solution by using it as an initialization for gradient-based optimization of the original cost function, achieving improved accuracy by converging to a better local minimum.

% However, since the minimization results in a challenging high-dimensional non-convex optimization problem, the authors of~\cite{keller2020hydra} consider the Lorentz model, use \eqref{eq:hyperbolic_distance_arcosh} transform distances by the hyperbolic cosines and minimize instead $\sum_{i,j}\|-\langle\bbu_i,\bbu_j\rangle-\operatorname{cosh}(D_{i,j})\|^2$. Given the form of \eqref{eq:minkowski}, this problem is efficiently solved using spectral decompositions. The Hydra+ variant uses this embedding as an initialization and applies further optimization to minimize the original cost function, yielding improved accuracy by converging to a local minimum.

% These methods address the problem of embedding data into hyperbolic space by minimizing a stress functional that measures the discrepancy between observed pairwise distances and hyperbolic distances in the embedding. Since stress minimization is highly non-convex, Hydra instead solves a relaxed problem by replacing the stress with a strain functional, which can be efficiently minimized using spectral decompositions. The Hydra+ variant uses this embedding as an initialization and applies further optimization to minimize the original stress functional, yielding improved accuracy by converging to a local minimum.

\noindent \textbf{Poincar\'e Embeddings \cite{nickel2017poincare}.} 
In this method, embeddings are learned in the Poincar\'e ball model. As we mentioned before, this model is conformal, allowing Riemannian gradients to be computed simply as Euclidean gradients rescaled by a position-dependent factor induced by the metric. After each update, embeddings are projected back into the unit ball to ensure validity of the representation.
Learning is thus performed using Riemannian stochastic gradient descent, minimizing an application-dependent loss function defined in terms of hyperbolic distances. In the case of network embedding, the probability of observing an edge between two nodes is modeled using a Fermi--Dirac distribution that decreases with their hyperbolic distance, and training proceeds by minimizing a cross-entropy loss with negative sampling. Such distribution is known to generate networks that exhibit real-life properties; e.g., strong clustering, small diameter, and heterogeneous degree distributions.

% This approach uses the fact that, as discussed before, the Poincar\'e ball model is conformal, so that gradients in the manifold are Euclidean gradients simply re-scaled by a position-dependent factor induced by the metric. The proposed method is thus a Riemannian stochastic gradient descent using this model, minimizing an application-dependent loss function. For instance, in our network embedding setting, a Fermi-Dirac distribution is used to model the probability of an edge as a decreasing function of the hyperbolic distance, and the objective function is a cross-entropy loss with negative sampling. 

% This approach uses the Poincaré ball model, which is a conformal model of hyperbolic space, so that the Riemannian gradient can be computed by appropriately rescaling the Euclidean gradient. The loss function is chosen depending on the application and typically depends on the hyperbolic distances between embeddings. In the network embedding setting, a Fermi-Dirac distribution is used to model the probability of an edge as a decreasing function of the hyperbolic distance; the objective function is a cross-entropy loss with negative sampling, and the optimization is carried out via stochastic projected gradient descent.

\noindent \textbf{Lorentz Embeddings \cite{nickel2018learning}.}
Proposed as an alternative to Poincar\'e embeddings, this approach performs representation learning in the Lorentz model instead, which the authors argue provides improved numerical stability for Riemannian optimization, particularly near the boundary of the Poincar\'e ball. 
Similarly to Hydra, this method assumes the availability of pairwise similarity information between nodes.
% The method assumes the availability of pairwise similarity information between nodes, which may be given explicitly (e.g., real-valued similarity scores) or implicitly through graph structure. 
Embeddings are learned so that more similar nodes are placed closer in hyperbolic space, using a softmax-based ranking loss that attracts similar pairs while pushing apart dissimilar ones. In the graph setting, direct neighbors are treated as the most similar nodes, while randomly sampled non-adjacent nodes serve as dissimilar examples.

\noindent \textbf{Mercator and D-Mercator  \cite{garcia2019mercator,jankowski2023dmercator}.} 
These methods operate in the native (polar) representation and assume the same underlying generative model as Poincar\'e Embeddings.
% in which the probability of connection also follows a Fermi--Dirac distribution that decreases with hyperbolic distance. Such models are known to generate networks that simultaneously exhibit strong clustering, small-world properties, and heterogeneous degree distributions. 
As discussed previously for the native model, radial coordinates can be interpreted as latent ``popularity'' variables controlling expected node degrees, while angular coordinates encode ``similarity'' between nodes. Mercator first computes an initial embedding by applying a model-corrected Laplacian Eigenmaps procedure to recover an approximate angular ordering of the nodes. This initial configuration is subsequently refined by maximizing the likelihood of the observed network under the assumed hyperbolic generative model. D-Mercator extends this framework by allowing similarity to be represented using $D$ angular dimensions. %, enabling the embedding of networks with more complex similarity structures.

\noindent \textbf{HyperMap \cite{papadopoulos2015networka}.}
This method relies on the same popularity--similarity generative model as (D-)Mercator, but explicitly assumes a chronological growth process of the network. Starting from the observed topology, HyperMap first infers an approximate arrival order of the nodes, using node degrees as proxies for their age under the growth model. Conditional on this inferred ordering, hyperbolic coordinates are assigned sequentially by maximizing, for each newly added node, a likelihood function derived from the model’s distance-dependent connection probability, using discrete sampling over angular positions. 
% The method requires as input the model parameters controlling average degree growth and clustering. 
Finally, a correction step is applied in which previously assigned angular coordinates are iteratively refined to improve consistency between the inferred geometry and the observed network topology.

% This method uses the same popularity/similarity model as (D-)Mercator and assumes a chronological growth of the network. Given a network adjacency matrix, HyperMap first infers the arrival order of nodes, using their degrees as proxies for node age. Conditional on this ordering, the hyperbolic coordinates of each node are then inferred by maximizing a local likelihood function through discrete sampling over angular positions. The method requires as input the model parameters controlling average degree growth and clustering, and includes a correction step in which previously assigned angular coordinates are refined to improve likelihood consistency with the observed topology.

\noindent \textbf{Poincar\'e Maps \cite{klimovskaia2020poincare}.} 
% This method was developed specifically for single-cell RNA sequencing data and begins by constructing a $k$-nearest-neighbor graph with Gaussian edge weights. 
Given a graph, and in particular its Laplacian matrix $\bbL$, the Relative Forest Accessibility (RFA) matrix is computed as $(\bbI+\bbL)^{-1}$. This matrix is doubly stochastic and can be interpreted as defining probability distributions over nodes. 
The embedding objective in Poincar\'e Maps is formulated by defining similarities between hyperbolic embeddings as normalized negative exponentials of the hyperbolic distance, and minimizing the Kullback--Leibler divergence between these similarities and the node-wise probability distributions induced by the RFA matrix. The resulting optimization problem is solved using Riemannian stochastic gradient descent in the Poincaré ball.

% Similarities between the sought embeddings are defined as a normalized negative exponential of the hyperbolic distance, and the Kullback-Leibler divergence between these similarities and the probabilities induced by the RFA matrix is minimized. The optimization is carried out using Riemannian stochastic gradient descent.

\section{Experiments}\label{sec:experiments}

% \red{Dar la mayor parte del peso acá a esta sección. Poner también tiempos de inferencia, aunque no sé si tenemos algo sobre complejidad, porque sería tiempos de nuestra implementación. Lo otro es si tuviéramos varias tiradas para mirar variación alrededor de cada resultado. Y por último, podríamos jugar con la dimensión del RDPG a ver si con $d$ anda mejor (o no).}

% All experiments are conducted using HypeGRL, ensuring consistent training, evaluation, and visualization across embedding methods.

\subsection{Hyperbolic Embedding Evaluation}\label{subsec:embedding_eval}

\begin{figure}
    \centering
    \includegraphics[width=1.0\linewidth]{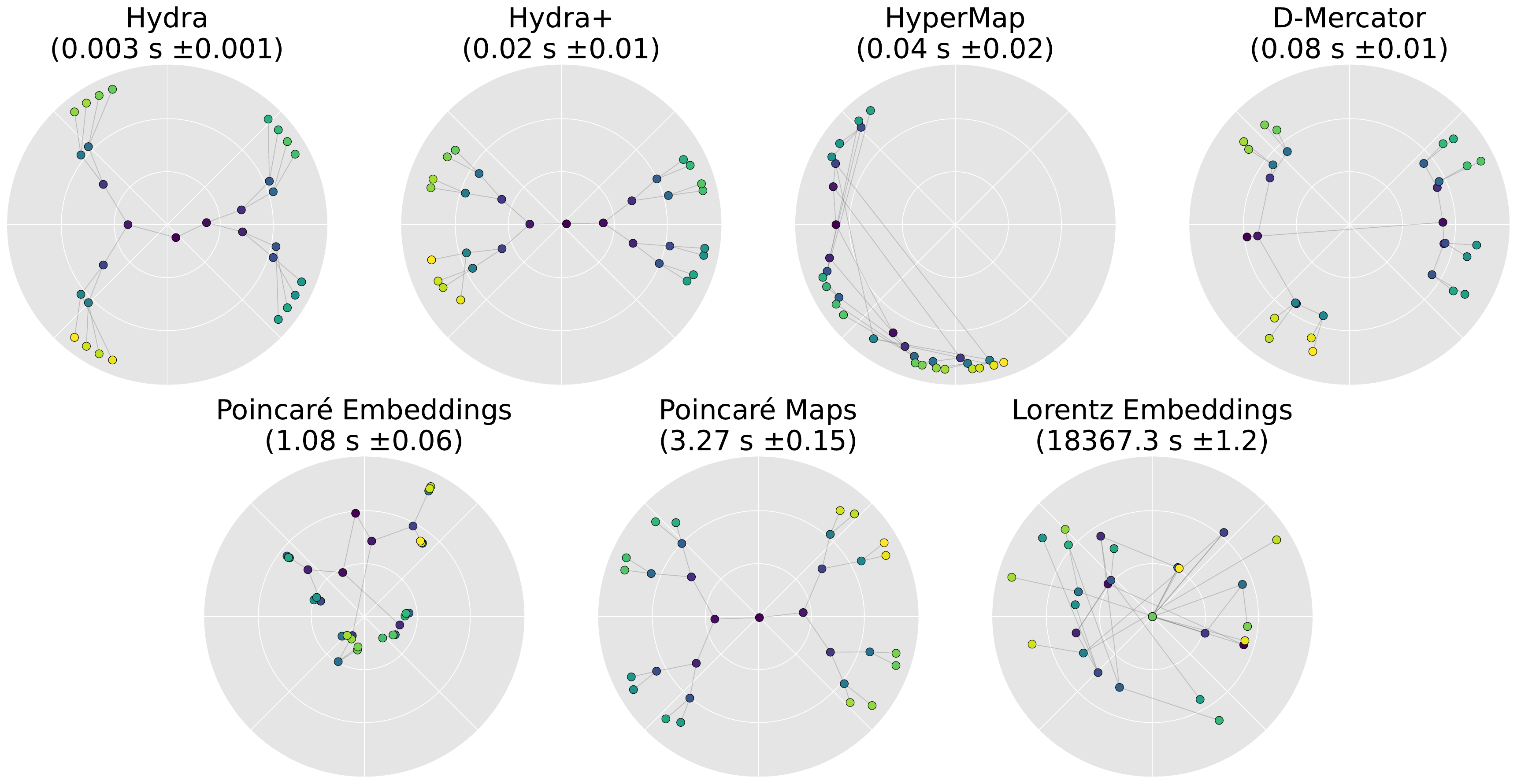}
    \caption{Comparison of hyperbolic embedding methods in the native representation (polar coordinates $r$, $\theta$) for a balanced binary tree (branching factor 2, depth 4, 31 nodes). 
% Embeddings are shown in the native coordinates ($r$, $\theta$). 
%, with nodes colored by depth. 
Mean execution time of 10 runs are indicated (the Lorentz model was run 3 times only). }
    \label{fig:embedding_tree_test}
\end{figure}

% We first evaluate hyperbolic embedding methods on a synthetic benchmark consisting of a balanced binary tree with branching factor 2 and depth 4, resulting in 31 nodes. This controlled hierarchical structure allows us to assess whether embeddings preserve the underlying tree geometry.
We begin with a simple and controlled setting consisting of a balanced binary tree with branching factor~2 and depth~4, yielding a total of~31 nodes. This hierarchical structure allows us to assess whether different embedding methods preserve the underlying tree geometry.
Results for all methods implemented in \texttt{HypeGRL}, using a two-dimensional hyperbolic space ($n=2$), are shown in Fig.~\ref{fig:embedding_tree_test}, visualized in polar coordinates $(r,\theta)$. As discussed previously, the radial coordinate reflects hierarchical depth, while the angular coordinate represents similarity between nodes. To facilitate visual inspection, nodes are colored according to their depth in the tree, with lighter colors indicating deeper levels, and the edges of the original tree are overlaid. The methods are ordered by their mean runtime, which is reported below each method's name.

% Let us first consider a relatively simple scenario, where we observe a balanced binary tree with branching factor 2 and depth 4, resulting in 31 nodes. This controlled hierarchical structure allows us to assess whether the embeddings methods preserve the underlying tree geometry. Results for all methods implemented in \texttt{HyperGRL} (using dimension $n=2$) are shown in Fig.\ \ref{fig:embedding_tree_test} visualized in polar coordinates ($r$, $\theta$). Recall that the radial coordinate reflects hierarchical depth and the angular coordinate represents similarity. To ease the visualization, lighter colors indicate a larger depth in the tree and edges are shown. Methods are sorted according to their mean execution time, which is also shown below the corresponding method's name.

% Most methods successfully recover the tree structure by assigning a smaller radius to higher-level nodes and a larger radius to leaf nodes. In particular, D-Mercator, Hydra, Hydra+, and Poincaré Embeddings produce clear hierarchical layouts that closely reflect the underlying tree structure. In contrast, HyperMap fails to consistently separate hierarchical levels.

Most methods successfully recover the hierarchical structure by assigning smaller radii to higher-level nodes and larger radii to leaf nodes, while clearly separating different branches of the tree in the angular dimension. In particular, Hydra, Hydra+, D-Mercator, and Poincar\'e Maps produce embeddings that closely reflect the underlying tree organization. Among them, Hydra is by far the fastest method, as its computation relies on a single eigen-decomposition of a transformed distance matrix. Although Hydra+ incurs an increase in runtime due to its additional optimization step, it remains highly competitive in this regard (along with HyperMap and D-Mercator) and consistently yields improved embeddings. 

% Most methods successfully recover the tree structure by assigning a smaller radius to higher-level nodes and a larger radius to leaf nodes, clearly separating the branches of the tree. In particular, Hydra, Hydra+, D-Mercator, and Poincar\'e Maps produce clear hierarchical layouts that closely reflect the underlying tree structure. In particular, the first method is by far the fastest, owing to its computation being based on a simple eigen-decomposition of a distance matrix. Although Hydra+ comes with a penalty in computation times, it is still extremely competitive in this terms, effectively producing improved embeddings. 
% In addition to embedding quality, we evaluate computational efficiency. Table \ref{tab:embedding_runtimes} reports the mean execution time over ten runs for each method. While most approaches complete within fractions of a second, Lorentz embeddings exhibit prohibitively large runtimes, rendering them impractical for further evaluation.

% Based on both embedding quality and runtime performance, we select D-Mercator, Hydra+, and Poincaré Embeddings as the most effective methods and evaluate them in the downstream experiments.

This experiment also provides useful insights into the limitations of several approaches. First, the default implementation of Lorentz embeddings for graphs exhibits prohibitively large computation times in this setting. These stem from the use of random neighbor sampling to define similarity pairs for computing the loss, suggesting that more efficient sampling strategies would be required for practical applicability. Second, HyperMap struggles to consistently separate hierarchical levels in this regular tree structure, as its assumption that node degree serves as a proxy for node age is ill-suited to graphs with uniform degree. For these reasons we focus on Hydra+, D-Mercator, Poincar\'e Embeddings and Poincar\'e Maps in the sequel.

% Some interesting insights can be obtained from this simple experiment. Firstly, the default implementation of Lorentz embeddings for graphs is simply impractical. Its high computation times stem from randomly choosing neighbors to measure (dis)similarity and computing the loss. Better sampling strategies need to be designed for the algorithm to be viable. Secondly, HyperMap fails to consistently separate hierarchical levels. Its assumption that a node's degree can be used as a proxy for its age in the underlying growth model clearly complicates the problem in a regular tree. Finally, although Poincar\'e Maps do produce reasonable embeddings, it tends to place nodes belonging to different branches very near each other. This, together with computation times orders of magnitude larger than most of the other methods, is the reason we will focus on D-Mercator, Hydra+, and Poincaré Embeddings as the most effective methods and evaluate them in the downstream experiments.

% \begin{table}[h]
% \centering
% \caption{Execution Time (10 runs per embedding type)}
% \begin{tabular}{l c}
% \hline
% \textbf{Embedding Type} & \textbf{Mean $\pm$ Std (s)} \\
% \hline
% Hydra                    & 0.00 $\pm$ 0.00 \\
% Hydra+              & 0.02 $\pm$ 0.01 \\
% HyperMap                 & 0.04 $\pm$ 0.02 \\
% D-Mercator                & 0.08 $\pm$ 0.01 \\
% Poincare Embeddings     & 1.08 $\pm$ 0.06 \\
% Poincare Maps           & 3.27 $\pm$ 0.15 \\
% Lorentz                  & 18367.3 $\pm$ 1.15 (*)\\
% \hline
% \end{tabular}
% \label{tab:embedding_runtimes}
% \end{table}

\subsection{Link prediction}\label{subsec:link_prediction}

Although visualization is an interesting task in its own right, embeddings are typically used for a downstream task. We now evaluate their their effectiveness for link prediction.
To this end, we use a standard edge removal protocol. 
Given an input graph $G=(V,E)$, each edge is independently retained with probability $q=0.9$. This results in three disjoint sets: $\Omega_E$, the retained edges used for training; $\Omega_R$, the removed edges serving as ground-truth positive examples; and $\Omega_N$, the set of true non-links corresponding to node pairs that were never connected in the original graph.
Embeddings are learned exclusively on the graph $G'=(V,\Omega_E)$, which preserves all nodes while restricting the observed edge set.
% We evaluate link prediction performance using a standard edge removal protocol. Given an input graph $G=(V,E)$, each edge is independently retained with probability $q = 0.9$, resulting in three disjoint sets: $\Omega_E$, the retained edges used for training; $\Omega_R$, the removed edges serving as ground-truth positive examples; and $\Omega_N$, the set of true non-links corresponding to node pairs that were never connected in the original graph. Embeddings are trained exclusively on the graph $G'=(V,\Omega_E)$, which preserves all nodes while restricting the observed edge set.

% For the three hyperbolic embedding methods considered here, learned embeddings are converted to hyperboloid coordinates and pairwise hyperbolic distances are computed. Candidate links in $\Omega_R \cup \Omega_N$ are ranked by increasing distance, with smaller distances indicating higher likelihood of connection. As a Euclidean baseline, we implement Random Dot Product Graph (RDPG) embeddings using adjacency spectral embedding, where links are predicted from the probability matrix $\mathbf P = \mathbf X \mathbf X^\top$ and ranked in descending order of connection probability.

To evaluate the three hyperbolic embedding methods considered here, we compute all pairwise hyperbolic distances and rank candidate links in $\Omega_R \cup \Omega_N$ accordingly, with smaller distances indicating a higher likelihood of connection.
As an Euclidean baseline, we estimate the aforementioned RDPG embeddings via adjacency spectral embedding, considering multiple embedding dimensions $n$.
In this case, links are predicted from the estimated probability matrix $\hbP$ and ranked in descending order of connection probability.

% Based on the embedding quality and runtime analysis in Section~\ref{subsec:embedding_eval}, we restrict the hyperbolic methods considered in this experiment to D-Mercator, Hydra+, and Poincar\'e Embeddings.

% To evaluate link prediction performance, we report the F1-score when predicting exactly $|\Omega_R|$ links. Since this is typically a very challenging task, we also report the lift metrics. These measure the concentration of true positive links among the highest-ranked predictions by comparing the number of correctly predicted links within a given fraction of the ranked list to a uniform random baseline. Higher lift values indicate that true links are more effectively concentrated near the top of the ranking. Experiments are conducted on three biological network datasets: Toggle Switch~\cite{toggle_switch_ref}, Olsson~\cite{olsson_ref}, and Myeloid Progenitors~\cite{myeloid_ref}, which are commonly used benchmarks for gene regulatory and cellular interaction networks. Results are reported in Table \ref{tab:link_prediction}.
To evaluate link prediction performance, we report the F1-score obtained when predicting exactly $|\Omega_R|$ links. Since this setting is particularly challenging, we additionally report lift metrics: the proportion of true positive links among the top 10\% highest-ranked predictions in $\Omega_R\cup \Omega_N$.
Higher lift values indicate that the corresponding distance is indeed informative for the link prediction task.
Experiments are conducted on three biological network datasets: Toggle Switch~\cite{toggle_switch_ref}, Olsson~\cite{olsson_ref}, and Myeloid Progenitors~\cite{myeloid_ref}, which are commonly used benchmarks for gene regulatory and cellular interaction networks. Table \ref{tab:link_prediction} reports results (bold and underline indicate best and second-best results respectively) and a few statistics about each network. 
Of particular interest is $\delta_{\text{mean}}$, which denotes the mean Gromov hyperbolicity of the graph. A small value relative to the graph diameter indicates that the graph exhibits pronounced hyperbolic structure~\cite{gilbert2024hyperbolicity}.

\begin{table}
\centering
\caption{Link Prediction results} 
% (10\% Edges Removed)}
\small
\resizebox{\columnwidth}{!}{
\begin{tabular}{c l c c}
\hline
\textbf{Dataset} & \textbf{Model} 
& \textbf{F1 Score (\%)} 
& \textbf{Lift (1st Decile)} \\
\hline

\multirow{6}{*}{
    \begin{tabular}{c}
        \textbf{Toggle Switch}\\ 
        ($|V|=200$, $|E|=1896$,\\
        $|\Omega_N|=37904$, \\
        Diam$=16$, $\delta_{\text{mean}}=0.16$)
        % hyperbolicity_mean:  0.161  hyperbolicity max:  1.5
        % {'delta_max': 1.5, 'delta_mean': 0.1623, 'delta_std': 0.25059670787941324}. normalized hyperbolicity: 0.09375
    \end{tabular}
    } 
 & RDPG ($n=2$)            
 & $13.0 \pm 2.0$ 
 & 62/189  \\
 % & RDPG ($n=4$)            
 % & $18.6 \pm 1.8$ 
 % & 121/189 \\
 & RDPG ($n=8$)            
 & $49.9 \pm 2.7$ 
 & 180/189 \\
 & \textbf{RDPG ($n=16$) }           
 & $\mathbf{74.8 \pm 2.6}$
 & \textbf{189/189} \\
 & Hydra+            
 & $51.6 \pm 1.8$ 
 & $\underline{187/189}$ \\
 & D-Mercator            
 & $\underline{57.7 \pm 3.4}$ 
 & 185/189 \\
 & Poincar\'e Embeddings 
 & $41.9 \pm 5.1$ 
 & 184/189 \\
 & Poincar\'e Maps 
 & $\underline{57.5 \pm 4.2}$
 & 185/189 \\
\hline

\multirow{6}{*}{
    \begin{tabular}{c}
        \textbf{Olsson}\\ 
        ($|V|=382$, $|E|=4214$,\\
        $|\Omega_N|=141328$,\\
        Diam$=8$, $\delta_{\text{mean}}=0.25$)
        %hyperbolicity_mean:  0.24933  hyperbolicity max:  1.5
        % {'delta_max': 1.5, 'delta_mean': 0.25028, 'delta_std': 0.2924891820221733}. normalized hyperbolicity: 0.1875
    \end{tabular}
} 
 & RDPG ($n=2$)            
 & $8.48 \pm 1.3$ 
 & 169/420 \\
 % & RDPG ($n=4$)            
 % & $13.5 \pm 1.4$ 
 % & 234/420 \\
 & RDPG ($n=8$)            
 & $\underline{19.0 \pm 1.6}$
 & 354/420 \\
 & \textbf{RDPG ($n=16$)  }          
 & $\mathbf{30.4 \pm 1.8}$
 & \textbf{390/420 }\\
 & Hydra+            
 & $8.67 \pm 1.7$ 
 & 305/420 \\
 & D-Mercator            
 & $15.1 \pm 1.8$ 
 & 328/420 \\
 & Poincar\'e Embeddings
 & $10.8 \pm 1.5$ 
 & 290/420 \\
 & Poincar\'e Maps 
 & $\mathbf{29.25 \pm 1.21}$
 &  \underline{387/420}\\
\hline

\multirow{6}{*}{
    \begin{tabular}{c}
        \textbf{Myeloid Progenitors}\\ 
        ($|V|=640$, $|E|=5649$,\\
        $|\Omega_N|=403311$,\\
        Diam$=38$, $\delta_{\text{mean}}=0.15$)
        % hyperbolicity_mean:  0.14555  hyperbolicity max:  1.5
        % {'delta_max': 1.5, 'delta_mean': 0.14408, 'delta_std': 0.23184251896492156}. normalized hyperbolicity: 0.039473684210526314
    \end{tabular}
} 
 & RDPG ($n=2$)            
 & $7.37 \pm 1.5$ 
 & 196/547 \\
 % & RDPG ($n=4$)            
 % & $11.1 \pm 1.0$ 
 % & 252/547 \\
 & RDPG ($n=8$)            
 & $15.0 \pm 1.3$ 
 & 397/547 \\
 & RDPG ($n=16$)            
 & $21.7 \pm 1.3$
 & \textbf{547/547} \\
 & Hydra+            
 & $22.0 \pm 1.6$ 
 & 219/547 \\
 & D-Mercator            
 & $\underline{41.9 \pm 3.1}$ 
 & $\underline{532/547}$ \\
 & Poincar\'e Embeddings 
 & $20.7 \pm 1.6$ 
 & 490/547 \\
 & \textbf{Poincar\'e Maps }
 & $\mathbf{48.4 \pm 6.2}$
 & \textbf{547/547} \\
\hline

\end{tabular}
}
\label{tab:link_prediction}
\end{table}

Notably, the majority of true links are concentrated within the first decile of the ranked lists produced by hyperbolic embeddings, indicating strong discriminative power.
In contrast, RDPG predictions using the same dimension as the hyperbolic methods ($n=2$) are more evenly distributed across deciles, suggesting weaker prioritization of true links among top-ranked candidates.
Among the hyperbolic methods considered, Poincar\'e Maps consistently achieves the best F1-score, followed by D-Mercator.
When compared with RDPG embeddings, hyperbolic approaches remain competitive despite using only two-dimensional representations.
% In the Olsson dataset, which is precisely the ``least hyperbolic'' of the three datasets according to $\delta_{\text{mean}}$, RDPG surpasses D-Mercator only when the embedding dimension is increased to $n=8$, and still performs worse than Poincar\'e Maps.
In the Myeloid dataset, which  is precisely the ``most hyperbolic'' of the three datasets according to $\delta_{\text{mean}}$, RDPG is always outperformed, even for $n=16$. When using the still high $n=8$, it outperforms most hyperbolic methods precisely in the Olsson dataset, the ``least hyperbolic'' of the three.
These results suggest that RDPG embeddings require substantially higher dimensionality to achieve performance comparable to that of low-dimensional hyperbolic embeddings.
% Moreover, in the Myeloid Progenitors dataset, hyperbolic embeddings, particularly D-Mercator, outperform RDPG even at higher Euclidean dimensions.
Overall, strongly hyperbolic graphs benefit significantly from low-dimensional hyperbolic representations.

% Overall, these results demonstrate the representational efficiency of hyperbolic embeddings, which achieve competitive or superior link prediction performance using low-dimensional spaces ($d=2$), whereas Euclidean baselines require substantially higher dimensionality.

\subsection{Node classification}\label{subsec:node_classification}

% We evaluate node classification performance by learning hyperbolic graph embeddings and performing classification with a $k$-nearest neighbors (KNN) classifier based on hyperbolic distance metrics. Node embeddings are obtained using Poincaré Embeddings and Hydra+.

% Classification is performed using KNN with the Poincar\'e distance, which measures the hyperbolic distance between points in the disk. This distance replaces the standard Euclidean metric commonly used in KNN classifiers. The classifier is trained on a subset of labeled nodes and evaluated on held-out test nodes. Performance is reported in terms of the F1-score for different values of $k$.

We evaluate node classification performance by first computing the embeddings and then applying a (hyperbolic) distance-based classifier in the embedding space. Our goal is to assess how well different hyperbolic embedding methods organize nodes of similar labels in low-dimensional hyperbolic space. Embeddings are thus computed using exclusively the graph topology and ignoring node labels. Classification is then performed using a $k$-nearest neighbors (KNN) classifier, where distances between nodes are measured using the hyperbolic metric associated with the embedding model. 
This choice intentionally avoids introducing an additional learned classifier and instead directly probes the geometric organization induced by each embedding method.
In all cases, embeddings are computed on the full graph, and labels are split into 80\% for training and 20\% for testing using stratified sampling. 
% (Poincaré ball or Hyperboloid). 

% We evaluate this procedure on the NeuroSEED\cite{polblogs_ref} and Polblogs\cite{neuroseed_ref} datasets. Results are summarized in Table~\ref{tab:node_classification}, together with previously reported results for HyperDT\cite{chlenski2024fast} and HyperRF\cite{chlenski2024fast}. These prior results serve as reference points for existing hyperbolic graph classification approaches.

% All experiments follow a transductive setting: embeddings are computed using the full graph structure, while labels are only used for training and evaluating the KNN classifier. 

% Performance is reported in terms of F1-score for different values of $k$. 
% 
We first consider a synthetic graph whose edges are generated from hyperbolic embeddings but do not strictly follow the generative models used in the methods under study. This setting allows us to evaluate robustness to model mismatch. In particular, we follow~\cite{chlenski2024fast}, and embeddings are generated using the wrapped normal distribution on the hyperboloid. We have $C=6$ classes, each with their randomly picked mean, and a graph is constructed by connecting each node to the other ten nearest nodes. 
Results are shown in Table \ref{tab:node_classification}.

\begin{table}
\centering
\caption{Node Classification Results}
% \red{Aclarar sobre cuántos nodos se hace la predicción (debería ser el mismo que en Hyper*). Agregar nº de classes?}}
\label{tab:hydra_hyper_comparison}
\resizebox{\columnwidth}{!}{
\begin{tabular}{l l c c c}
\hline
\textbf{Dataset} & \textbf{Model} & \textbf{$n$} & \textbf{F1 Score (\%)} \\
\hline
\multirow{6}{*}{
    \begin{tabular}{c}
        \textbf{Gaussian}\\ 
        ($|V|=1250$, $|E|=12580$, \\
        $C=6$)
        % ,\\
        % Diam$=x$, $\delta_{\text{mean}}=0.x$)
    \end{tabular}
}
 & \textbf{Hydra+ KNN ($k=5$) } & 2  & $\mathbf{85.03 \pm 0.19}$ \\
 & Hydra+ KNN ($k=10$) & 2  & $\underline{84.34 \pm 0.33}$ \\
 & D-Mercator KNN ($k=5$) & 2 & 17.81 $\pm$ 2.54 \\
 & D-Mercator KNN ($k=10$) & 2 & 17.32 $\pm$ 1.43\\
 & Poincaré Emb. KNN ($k=5$)  & 2  & 77.16 $\pm$ 0.10 \\
 & Poincaré Emb. KNN ($k=10$) & 2  & 77.14 $\pm$  0.15 \\
 & Poincaré Maps KNN ($k=5$) & 2 & 83.04 $\pm$ 0.38 \\
 & Poincaré Maps KNN ($k=10$) & 2 & $\underline{83.92 \pm 0.10}$ \\
 % & HyperDT            & 2   & 39.98 $\pm$ 3.14 \\
 % & HyperRF            & 2   & 40.22 $\pm$ 3.35 \\
\hline
\multirow{6}{*}{
\begin{tabular}{c}
        \textbf{Polblogs}\\ 
        ($|V|=1222$, $|E|=16717$, \\
        $C=2$)
        % ,\\
        % Diam$=x$, $\delta_{\text{mean}}=0.x$)
    \end{tabular}
}
 & Hydra Plus KNN ($k=5$)  & 2  & 77.54 $\pm$ 0.57 \\
 & Hydra Plus KNN ($k=10$) & 2 & 78.90 $\pm$ 0.38 \\
 & D-Mercator KNN ($k=5$) & 2 & 51.43 $\pm$ 3.53 \\
 & D-Mercator KNN ($k=10$) & 2 & 49.97 $\pm$ 3.97 \\
 & Poincaré Emb. KNN ($k=5$)  & 2  & $\underline{94.69 \pm 0.11}$ \\
 & \textbf{Poincaré Emb. KNN ($k=10$)} & 2  & $\mathbf{95.10 \pm 0.10}$ \\
 & Poincaré Maps KNN ($k=5$) & 2 & 80.26 $\pm$ 0.39 \\
 & Poincaré Maps KNN ($k=10$) & 2 & 80.81 $\pm$ 0.66 \\
 % & HyperDT            & 2  & 71.04 \textsuperscript{\cite{chlenski2024fast}} \\
 % & HyperRF            & 2  & 71.40 \textsuperscript{\cite{chlenski2024fast}} \\
\hline
\end{tabular}
}
\label{tab:node_classification}
\end{table}

% as follows. We construct a hyperbolic mixture of Gaussians, where each Gaussian component is defined using a wrapped normal distribution on the hyperboloid. The means of the Gaussian components are sampled from a normal distribution in the tangent plane at the origin and then projected onto the hyperboloid using the exponential map.

The first observation is that Hydra+ achieves the highest performance, which is consistent with its distance-based formulation. 
Furthermore, Poincar\'e Maps obtains very competitive results,  showing that the RFA approach used in this method is very flexible. On the other hand, D-Mercator performs very poorly in this case. Note furthermore that Poincar\'e Embeddings use a very similar underlying generative model but obtains significantly better results, suggesting that performance differences are largely driven by the estimation procedure rather than the underlying geometric model.
We confirm this trend on the real-world Polblogs dataset (lower part of Table \ref{tab:node_classification})~\cite{polblogs_ref}. In this case, Poincar\'e Embeddings obtains the best results, but D-Mercator still performs significantly below the rest of the methods.  We emphasize that the strong performance of KNN using some of the hyperbolic embedding methods does not imply superiority over more complex classifiers in general, but rather indicates that, for these datasets, class information is already well aligned with hyperbolic distances. 

\section{Conclusions}\label{sec:conlusiones}

In this work, we introduced \texttt{HypeGRL}, an open-source framework designed to unify the training, evaluation, and comparison of hyperbolic GRL methods. Through an experimental study across visualization, link prediction, and node classification tasks, we highlighted how different hyperbolic embedding approaches vary in terms of representation efficiency, computation cost, and downstream performance. 
Our results show that Poincar\'e Maps is overall the best performing method in all tasks, albeit at the expense of significantly higher computation times. 
% Hydra+ offers an interesting balance between performance and computation cost.  
In contrast, Hydra+ emerges as a compelling alternative, offering competitive accuracy while maintaining significantly lower computational cost, and thus providing a favorable balance between performance and efficiency.
% Hydra+, Poincar\'e Embeddings and D-Mercator can efficiently capture meaningful graph structure and support strong predictive performance when combined with simple, geometry-aware decision rules. 
By providing a reproducible and extensible environment for evaluating hyperbolic embeddings, \texttt{HypeGRL} aims to facilitate further comparison, deeper understanding, and broader adoption of hyperbolic geometry in graph learning applications.

\bibliographystyle{IEEEtran}
\bibliography{refs}

\end{document}